\documentclass{sig-alternate}
\usepackage{times}
\usepackage{helvet}
\usepackage{courier}
\usepackage{algorithmic}
\usepackage{algorithm}
\usepackage{graphicx}
\usepackage{amsmath}
\usepackage{amsfonts}
\usepackage{bbm}
\usepackage{color}
\usepackage{multirow}
\usepackage{amssymb}
\usepackage{epstopdf}

\newtheorem{proposition}{Proposition}

\newcommand{\beq}{\begin{equation}}
\newcommand{\eeq}{\end{equation}}
\newcommand{\beqa}{\begin{eqnarray}}
\newcommand{\eeqa}{\end{eqnarray}}

\newcommand{\efig}{\end{figure}}

\newcommand{\mb}[1]{{\mathbf{#1}}}
\newcommand{\ie} {{\it i.e., }}
\newcommand{\eg} {{\it e.g., }}

\newcommand{\hide}[1]{}

\begin{document}
%

\title{Active Metric Learning from Relative Comparisons}
%
%
%
%
%

\numberofauthors{4} 
%
\author{
%
%
Sicheng Xiong$^\dag$ \qquad R\'{o}mer Rosales$^\ddag$ \qquad Yuanli Pei$^\dag$ \qquad Xiaoli Z. Fern$^\dag$ \\
\\
\affaddr{$^\dag$School of EECS, Oregon State University. Corvallis, OR 97331, USA}\\
\email{\{xiongsi, peiy, xfern\}@eecs.oregonstate.edu} \\
\\
\affaddr{$^\ddag$LinkedIn. Mountain View, CA 94043, USA} \\
\email{rrosales@linkedin.com}\\
}


\maketitle
\begin{abstract}

This work focuses on active learning of distance metrics from relative
comparison information. A relative comparison specifies, for a data
point triplet $(x_i,x_j,x_k)$, that instance $x_i$ is more similar to
$x_j$ than to $x_k$. Such constraints, when available, have been shown
to be useful toward defining appropriate distance metrics.  In
real-world applications, acquiring constraints often require
considerable human effort. This motivates us to study how to select
and query the most useful relative comparisons to achieve effective
metric learning with minimum user effort. Given an underlying class
concept that is employed by the user to provide such constraints, we
present an information-theoretic criterion that selects the triplet
whose answer leads to the highest expected gain in information about
the classes of a set of examples.  Directly applying the proposed
criterion requires examining $O(n^3)$ triplets with $n$ instances,
which is prohibitive even for datasets of moderate size. We show that
a randomized selection strategy can be used to reduce the selection
pool from $O(n^3)$ to $O(n)$, allowing us to scale up to larger-size
problems. Experiments show that the proposed method consistently
outperforms two baseline policies.

\end{abstract}

\category{}{}{}
\category{}{}{}[]

\terms{}


\section{Introduction}
\label{sec:intro}
Distance metrics play an important role in many machine learning algorithms. As such, distance metric learning has been heavily studied as a central machine learning and data mining problem. One category of distance learning algorithms \cite{rosales2006learning,schultz2003learning,xing2003distance} uses user-provided side information to introduce constraints or hints that are approximately consistent with the underlying distance. In this regard, Xing et al. \cite{xing2003distance} used pairwise information specifying whether two data instances are similar or dissimilar. Alternatively, other studies such as \cite{rosales2006learning,schultz2003learning} consider constraints introduced by relative comparisons described by \emph{triplets} $(x_i,x_j,x_k)$, specifying that instance or data point $x_i$ is more similar to instance $x_j$ than to instance $x_k$.


This paper addresses \emph{active learning from relative comparisons} for distance metric learning. We have chosen to focus on relative comparisons because we believe they are useful in a larger variety of contexts compared to pairwise constraints. Research in psychology has revealed that people are often inaccurate in making absolute judgments, but are much more reliable when judging comparatively \cite{nunnally94}. Labeling a pair of instances as either similar or dissimilar is an absolute judgment that requires a global view of the instances, which can be difficult to achieve by inspecting only a few instances. In contrast, relative comparisons allow the user to judge comparatively, making it more reliable and at the same time less demanding on the user.

While in some scenarios one can acquire relative comparisons
automatically (\eg using user click-through data in information
retrieval tasks \cite{schultz2003learning}), for many real
applications, acquiring such comparisons requires manual inspection of
the instances. This is very often time consuming and costly. For
example, in one of our applications, we want to learn a distance
metric that best categorizes different vocalizations of birds into
their species. Providing a relative comparison constraint in this case
will require the user to study the bird vocalizations either by
visually inspecting their spectrograms or listening to the audio
segments, which is very time consuming. This motivates us to study the
active learning problem to optimize the selection of relative
comparisons in order to achieve effective metric learning with minimum
user effort. Existing studies have only considered randomly or
sub-optimally selected triplets that are then provided to the metric
learning algorithm. To the best of our knowledge, there has been no
attempt to optimize the triplet selection for metric learning.



In this paper, we propose an information theoretic objective that selects a query whose answer will,
in expectation, lead to the most information gain. While the proposed objective can be used with any
metric learning algorithm that considers relative comparisons, a main obstacle is that we need to evaluate $O(n^3)$
triplets to select the optimizing query, where $n$ is the number of instances.
To reduce this complexity for selecting a triplet, we introduce a simple yet effective sampling procedure
that is guaranteed to identify quasi-optimal triplets in $O(n)$ with clear performance guarantees.
Experimental results show that the triplets selected by the proposed active learning method allow for:
1) learning better metrics than those selected by two different baseline policies, and
2) increasing the classification accuracy of nearest neighbor classifiers.




\section{Related Work}
\label{sec:rela}
\paragraph{Distance metric learning}
Xing et al.~proposed one of the first formal approaches for distance metric learning with side information \cite{xing2003distance}. In this study, they considered pairwise information indicating whether two instances are similar or dissimilar. A distance metric is learned by minimizing the distances between the instances in the similar pairs while keeping the distances between dissimilar pairs large.

Distance metric learning with relative comparisons has also been
studied in different contexts \cite{rosales2006learning,schultz2003learning}. Schultz et al.~formulated a constrained optimization problem where the
constraints are defined by relative comparisons and the objective is to learn a distance metric that remains as close to an un-weighted Euclidean metric as possible \cite{schultz2003learning}. Rosales et al.~\cite{rosales2006learning} proposed to learn a projection matrix from relative comparisons. This approach also employed relative comparisons to create constraints on the solution space, but optimized a different objective to encourage sparsity of the learned projection matrix. Both studies assumed that the relative comparisons are given \emph{a priori} and the constraints are a set of random or otherwise non-optimized set of pre-selected triplets. That is, the algorithm is not allowed to request comparisons outside of the given set.

\paragraph{Active learning}
There is a large body of literature on active learning for supervised classification~\cite{settles2010active}. One common strategy for
selecting a data instance to label is uncertainty sampling~\cite{lewis1994seq} where the instance with the highest label uncertainty is selected to be labeled. In Query-by-committee \cite{seung1992query}, multiple models (committee) are trained on different versions of the labeled data, and the
unlabeled instance with the largest disagreement among the committee members is queried for labeling. The underlying motivation is to efficiently reduce the uncertainty of the model. A similar motivation is also used in this paper. Other representative techniques include selecting the instance that is closet to the decision boundary (min margin) \cite{tong2002support}, and selecting the instance that leads to the largest expected error reduction \cite{roy2001toward}.

Active learning has also been studied for semi-supervised
clustering. In relation to our work, most previous approaches
concentrate on active selection of pairwise
constraints~\cite{basu2004active,gre2007cons,malla2008active,xu2005active}. While
the goal in these approaches is semi-supervised learning (not distance
metric learning), we partially share their motivation. In the context
of distance metric learning, an active learning strategy was proposed
in \cite{yang07} whithin the larger context of a Bayesian metric
learning formulation. This is the most closely related work as it
addresses active learning. However, like all of the above
formulations, it uses constraints of the form {\it must-link} and {\it
cannot-link}, specifying that two instances must or must not fall into
the same cluster respectively. As discussed previously, answering
pairwise queries as either must-link or cannot-link constraints
requires the user to make absolute judgements, making it less
practical/more demanding for the user and also more prone to human
error. In addition, none of the above formulations considers the
effect of {\it don't know} answers (to a triplet relationship query)
despite its importance in real, practical applications. These
relevant factors motivated us to study active learning in the current
setting. This is a problem that has not been studied previously.


\section{Proposed Method}
\label{sec:method}
The problem addressed in this paper is how to efficiently choose triplets to query in order to learn
an {\it appropriate} metric, where efficiency is defined by query complexity. More specifically, the
fewer queries/questions asked in order to achieve a specific performance the higher the efficiency.
A query is defined as a request to a user/oracle to label a triplet.
We view this as an iterative process, implying that the decision for query selection
should depend on what has been learned from all previous queries.

\begin{figure}[t!]
\begin{center}
\includegraphics[width=1.2in]{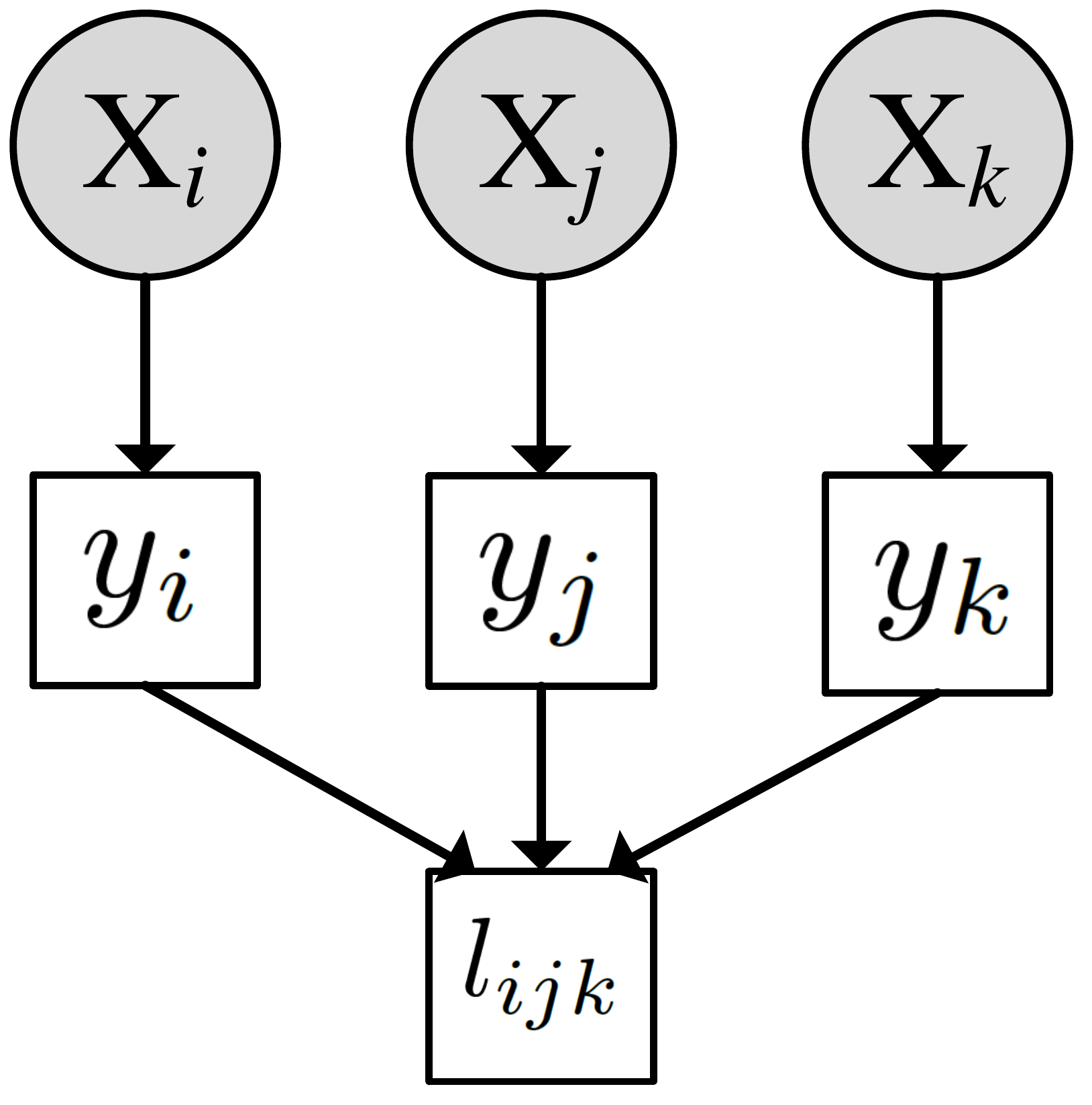}
\caption{Graphical model describing the relationships between three points $(\mb{x}_i,\mb{x}_j,\mb{x}_k)$, their class labels $(y_i, y_j, y_k)$, and the answer variable $l_{ijk}$. For simplicity we only show the parent-child relationships between three points $(i,j,k)$.
}
\label{fig:gm}
\end{center}
\end{figure}

\subsection{Problem Setup}
\label{sec:setup}

Given data $\mathcal{D} = \{\mathbf{x_1}, \cdots, \mathbf{x_n}\}$, we assume that there is an underlying
unknown class structure that assigns each instance to one of the $C$ classes. We denote the unknown labels
of the instances by ${\mathbf y} = \{ y_1, \cdots, y_n\}$, where each label $y_i \in {\mathcal Y} \triangleq \{1, \cdots, C\}$.
In the setting addressed, it is impossible to directly observe the class labels. Instead, the information about
labels can only be obtained through relative comparison queries.

A relative comparison query, denoted by triplet $r_{ijk} = (\mathbf{x}_i, \mathbf{x}_j, \mathbf{x}_k)$, can be interpreted as a question: ``{\it Is $\mathbf{x}_i$ more similar to $\mathbf{x}_j$ than $\mathbf{x}_k$?}''. Given a query $r_{ijk}$, a oracle/user will return an answer, denoted by $l_{ijk} \in {\cal A} \triangleq \{$yes, no, dk$\}$, based on the classes to which the three instances belong.
In particular, the oracle returns:
\begin{itemize}
\item $l_{ijk} = $ yes if $y_i=y_j\neq y_k$,
\vspace{-1em}
\item $l_{ijk} = $ no if $y_i\neq y_j = y_k$, and
\vspace{-1em}
\item $l_{ijk} = $ dk ({\it do not know}) for all other cases;
\end{itemize}
where the answers are expected to be noisy (as usual class labels are not guaranteed correct). This oracle is consistent with that used in prior work \cite{rosales2006learning,schultz2003learning}, except for one difference. Previous work only considered triplets that are provided \emph{a priori} and have either yes or no labels (by construction). In the setting addressed in this paper, the queries are selected by an active learning algorithm and must consider the possibility that the user cannot provide a yes/no answer to some queries.

We consider a {\it pool-based} active learning setting where in each iteration we have access to a {\it pool set} of triplets to choose from is available.
Let us denote the set of all labeled triplets by $R_l = \{(r_{ijk},l_{ijk}): l_{ijk} \,{\rm is} \,{\rm observed}\}$; likewise we denote the set of unlabeled triplets by $R_u$, which is also the pool set. In each active learning iteration, given inputs $\mathcal{D}$ and $R_l$, our active learning task is to optimally select one triplet $r_{ijk}$ from $R_u$ and query its label from the user/oracle. 

\subsection{A Probabilistic Model for Query Answers}
\label{sec:oracle}
A key challenge faced in an active learning problem lies in measuring how much {\it help} (an answer to) a query provides.
To answer this question, we model the relationship between data points, their class labels, and query answers in a probabilistic manner.



We denote the labels for all triplets that exist in $\mathcal{D}$ by $\mathbf{l}=\{l_{ijk}\}$ and let $l_{ijk}$ be random variables. We explicitly include the instance class labels $\mathbf{y}=\{ y_1, \cdots, y_n\}$ in our probabilistic model and assume that the query answer is independent of the data points given their class labels.
Formally, the conditional probability of the triplet labels and class labels is given by
\begin{equation}
\label{eq:pmodel}\nonumber
p(\mathbf{l},\mathbf{y}|\mathcal{D}) = \prod_{\{ijk: l_{ijk} \in \mb{l}\}} \hspace{-0.05in}{ p(l_{ijk}|y_i,y_j,y_k)
                                                \prod_{h \in \{i,j,k\}} \hspace{-0.05in} p(y_h|\mb{x}_h) }
\end{equation}

That is, query answer $l_{ijk}$ indirectly depends on the data points
through their unknown class labels $y_i$, $y_j$, and $y_k$, and the
class label of each data point only depends on the data point itself.
This set of independence assumptions is depicted by the graphical
model in Fig.\ref{fig:gm}, where the circles represent continuous
random variables and squares represent discrete random variables. In
addition, shaded nodes denote observed variables and unshaded ones
denote latent variables. For simplicity, the graphical model only
shows the relationships for one set of three points $(i,j,k)$. In
total, there are $O(n^3)$ query-answer random variables and each label
$y_i$ is the parent of $O(n^2)$ answer variables.

%
%

\subsection{Active Learning Criterion}
\label{sec:active}

In an active learning iteration, we seek to optimally select one triplet $r_{ijk}$ from a pool of unlabeled triplets $R_u$ and query its label from the oracle.  Our goal is to efficiently reduce the model uncertainty. To this end, we consider a selection criterion that measures the information that a triplet's answer provides about the class labels of the three involved instances. More specifically, we employ the mutual information (MI) function \cite{Cover91}. In our scenario, we use it to measure the degree upon which an answer to a triplet query $r_{ijk}$ reduces the uncertainty of the class labels $y_i$, $y_j$ and $y_k$.
As a measure of statistical dependence, we use MI to determine the expected reduction in class uncertainty achievable when observing the label of a triplet. This choice is suited for applications whose goal is classification since it targets the reduction of class uncertainty.

To formally define the above criterion, let $y_{ijk} = (y_i,y_j,y_k) \in {\cal Y}^3$. The unlabeled triplet $r_{ijk} \in R_u$ is chosen so that it maximizes the mutual information between the label of triplet $r_{ijk}$ (denoted by $l_{ijk}$) and the labels of three data points in $r_{ijk}$ (denoted by $y_{ijk}$) given $\mathcal{D}$ and $R_l$ (the labels of all previous queries).

In order to simplify the notation, we omit the term $\mathcal{D}$ in the probability expressions. It should be apparent from the context that the probabilities are conditioned on $\mathcal{D}$. With this simplification, our objective is given by:
\begin{eqnarray}
\label{eq:firstObj}
(ijk)^* &=& \underset{r_{ijk} \in R_u}{\arg\max} \ I (y_{ijk}; l_{ijk}|R_l) \\\nonumber
 &=& \underset{r_{ijk} \in R_u}{\arg\max} \ H(y_{ijk}|R_l) - H(y_{ijk}|l_{ijk}, R_l) \\ \nonumber
	&=&\underset{r_{ijk} \in R_u}{\arg\max}   \ H(y_{ijk}|R_l)  \\ \nonumber
	&& - \sum_{a \in {\cal A}} p (l_{ijk} = a|R_l) H(y_{ijk}|l_{ijk} = a, R_l) \\ \nonumber
	&=&\underset{r_{ijk} \in R_u}{\arg\max} \ (1-p (l_{ijk} = {\rm dk}|R_l)) H(y_{ijk}|R_l) \\ \nonumber
	& & -\sum_{a \in \{\rm yes,no\}} p (l_{ijk} = a|R_l)  H(y_{ijk}|l_{ijk} = a, R_l)
\end{eqnarray}

The last step of the above derivation assumes that a dk ({\it don't know}) answer provides no information to the metric learning algorithm, \ie $H(y_{ijk}|l_{ijk} = {\rm dk}, R_l) = H(y_{ijk}|R_l)$. This assumption is not required by the proposed approach, but it is employed to  address the fact that dk answers are not considered by existing distance metric learning agolrithms. This modeling limitation in existing metric learning algorithms does not affect batch (non-active) learning approaches.
In information theoretic terms, this need not to be the case. Future metric learning algorithms may model this situation differently. However, the {\it off-the-shelf} metric learning algorithm we use in this paper does not explicitly model such situations\footnote{This is largely an open research question beyond the scope of this paper.}.

Eq.~\ref{eq:firstObj} selects the triplet with the highest return of information about the class membership of the involved instances. In particular, focusing on the first term of Eq.~\ref{eq:firstObj}, we note that it will avoid selecting a triplet if it has a high probability of returning dk. Furthermore, it will also avoid selecting a triplet whose class uncertainty is low, which is consistent with a commonly used active learning heuristic that selects queries of high uncertainty. The second term in the equation helps with avoiding triplets whose yes or no answer provide little or no help in resolving the class uncertainty.

To compute Eq.~\ref{eq:firstObj}, two terms need to be specified, $H(y_{ijk}|R_l)$ and $H(y_{ijk}|l_{ijk}=a, R_l)$ for $a\in\{$yes, no$\}$.
First, we apply our model independence assumption that labels are conditional independent from each other given the data points and the
labeled triplets $R_l$ and rewrite $H(y_{ijk}|R_l)$ as:
\begin{equation}
\label{eq:yijk}
\begin{array}{rl}
H(y_{ijk}|R_l) =&\hspace{-0.1in} \displaystyle \sum_{h \in \{i, j, k\}} \hspace{-0.05in}H(y_h|R_l) \\
                                =&\displaystyle  - \hspace{-0.1in} \sum_{h \in \{i, j, k\}}\sum_{c=1}^C p(y_h = c|R_l) \log p(y_h = c|R_l)
\end{array}
\end{equation}

To compute $H(y_{ijk}|l_{ijk} = a, R_l), a \in \{$yes, no$\}$, applying the definition of
entropy and Bayes Theorem, it is easy to show that:

\begin{eqnarray}
\label{eq:yndk}
\nonumber
&&H(y_{ijk}|l_{ijk} = a, R_l) \\ \nonumber
& &= -\sum_{y_{ijk} \in {\cal Y}^3} p(y_{ijk}|l_{ijk} = a, R_l) \log p(y_{ijk}|l_{ijk} = a, R_l)\\ \nonumber
& &=-\sum_{y_{ijk} \in {\cal Y}^3} \frac{p(l_{ijk}=a|y_{ijk})p(y_{ijk}|R_l)}{p(l_{ijk}=a|R_l)} \\  \nonumber
& &\;\;\;\;\; \times \log \frac{p(l_{ijk}=a|y_{ijk})p(y_{ijk}|R_l)}{p(l_{ijk}=a|R_l)} \\
\end{eqnarray}
There are three key terms in Eq~\ref{eq:yndk}.  The first term $p(l_{ijk}=a|y_{ijk})$ is a deterministic distribution assigning probability one to whichever value $l_{ijk}$ should take according to the oracle based on $y_{ijk}$. For example, if $y_i=y_j\neq y_k$, we have $p(l_{ijk} = $yes$|y_{ijk})=1$.
By the independence assumption, the second term $p(y_{ijk}|R_l)$ can be factorized as
$\underset{h=i,j,k}{\prod}p(y_{h}|R_l)$

The last term is $p(l_{ijk}=a|R_l)$. Note that $p(y_h|R_l)$ and $p(l_{ijk}=a|R_l)$ are the only unspecified quantities for computing the objective in Eq.~\ref{eq:firstObj} and will need to be estimated from data. We will delay the discussion of how to estimate these probabilities until Section~\ref{sec:pest} so that we can complete the high level description of the
algorithm. For now, we will assume there exists a method to estimate these probabilities and proceed to discuss how to use the proposed selection criterion.

\subsection{Scaling to Large Datasets}
\label{sec:prune}

Based on the formulation thus far,  at each iteration the active learning algorithm works by selecting a triplet from the set of all unlabeled triplets
$R_u$ that maximizes the above introduced objective Eq. \ref{eq:firstObj}.
In the worst case, there are $O(n^3)$ triplets in $R_u$ for a data set of $n$ points.
To reduce this complexity, we propose to construct a smaller pool set $R_p$ by randomly sampling from $R_u$. We then select a triplet that maximizes the selection criterion from $R_p$.
Therefore, our objective function becomes
\begin{eqnarray}
\label{eq:fiOb}
(ijk)^* &=& \underset{r_{ijk} \in R_p}{\arg\max} \ I (y_{ijk}; l_{ijk}|R_l) 
\end{eqnarray}

Although the maximizing triplet in $R_p$ might not be the optimal triplet in $R_u$,
we argue that this will not necessarily degrade the active learning performance by much. Specifically, we expect significant redundancy among the unlabeled triplets. In most large datasets, a variety of triplets are near optimal and therefore, selecting a near-optimal triplet in $R_u$ (by choosing
the optimal triplet in $R_p$), instead of selecting the exact maximum, will not impact the performance significantly.
The above comes with a caveat as this approach is effective only if we can guarantee with high probability to select a triplet that is near optimal. Specifically, we consider a query to be near-optimal if it is $\epsilon$  top-ranked in $R_u$ according to the proposed selection criterion, where $\epsilon$ is a small number (e.g., $\epsilon = 0.01$).


It is indeed possible for the above approximation approach to guarantee a good triplet selection with high probability. We characterized this by the following proposition.
\begin{proposition}
When selecting the best triplet $r_{ijk}^*$ from $R_p$ according to Eq. \ref{eq:fiOb}, the probability that the selected $r_{ijk}^*$ is among the $\epsilon$ top ranked triplets of $R_u$ is $1 - (1-\epsilon)^{|R_p|}$, where $|R_p|$ denotes the cardinality of $R_p$.
\end{proposition}

The proposition can be easily proved by observing that the probability of obtaining a triplet not in the top $\epsilon$
set is $(1-\epsilon)$, and we are repeating the selection/sampling procedure $|R_p|$ times. Thus,  $1 - (1-\epsilon)^{|R_p|}$ gives us the probability of obtaining a near optimal query. 

In this work, we set $|R_p|$ to $100n$, where $n$ is the number of data points. This allows us to reduce the complexity from $O(n^3)$ to $O(n)$, and yet guarantee with high probability (greater than 99.99\% even for a small $n=100$) to select a triplet in the top $0.1\%$ of all unlabeled triplets.
We summarize the proposed active learning method in Algorithm \ref{alg:sum}.

\begin{algorithm}[t]
\caption{{\it Active Learning from Relative Comparisons}}
\textbf{Input}: data $\mathcal{D}=\{\mathbf{x}_1, \mathbf{x}_2, \cdots,\mathbf{x}_n\}$, the limit of queries and the oracle; \\
\textbf{Output}: a set of labeled triplets $R_l$; \\
\label{alg:sum}
\begin{algorithmic}[1]
   \STATE Initialize $R_l = \emptyset$. $R_u = $ the set of all triplets generated by $\mathcal{D}$.
   \STATE Generate the subset $R_p \subset R_u$ by random sampling.
   \REPEAT
   		\STATE Use the steps in Section \ref{item:ranForest} to estimate $p(y_i|R_l)$ for $i=1,\cdots,n$.
      \FOR{every triplet $r_{ijk} \in R_p$}
        \STATE Compute $p(l_{ijk}=a|R_l)$ for $a= $yes, no, dk using Eq.~\ref{eq:answer}
        \STATE Compute $I (y_{ijk}; l_{ijk}|R_l)$ using Eq.~\ref{eq:yijk}, \ref{eq:yndk}
      \ENDFOR
   \STATE $r_{ijk}^* = \arg\max_{r_{ijk} \in R_p} I (y_{ijk}; l_{ijk}|R_l)$
   \STATE Query the label of $r_{ijk}^*$ from the oracle.
   \STATE $R_l = R_l \bigcup r_{ijk}^*$, $R_p = R_p \backslash r_{ijk}^*$.
   \UNTIL{the limit of queries has been reached}
\end{algorithmic}
\end{algorithm}

\subsection{Probability Estimates and Implementation Details}
\label{sec:pest}

The above formulation requires estimating several probabilities. In particular, to evaluate the objective value achieved by
a given triplet $(i,j,k)$, we need to estimate $p(y_h=c|R_l)$ for $h\in\{i,j,k\}$ and $c\in \{1,...,C\}$, the probability
of each of the three data points belongs to each of the possible classes, and $p(l_{ijk}|R_l)$, the probability of different
answers to the given triplet, as analyzed in section \ref{sec:active}.
We will describe how to estimate these probabilities in this section.

First, we note that if we have $p(y_h=c|R_l)$ for $h\in\{i,j,k\}$ and $c\in \{1,...,C\}$, the probability of
different query answers can be easily computed as follows:
\begin{eqnarray}
\label{eq:answer}
\quad\quad p(l_{ijk} = {\rm yes}|R_l)=&&\hspace{-0.25in} \displaystyle \sum_{c=1}^C p(y_{i} = c|R_l)p(y_j = c|R_l)  \\
                                      &&\hspace{-0.25in} \times (1 -p(y_k = c|R_l) ) \nonumber
\end{eqnarray}
\begin{eqnarray*}
\quad\quad p(l_{ijk} = {\rm no}|R_l) = &&\hspace{-0.25in} \displaystyle \sum_{c=1}^C p(y_{i} = c|R_l)p(y_{k} = c|R_l) \\
                                       &&\hspace{-0.25in} \times (1 -p(y_j=c|R_l))
\end{eqnarray*}
\begin{eqnarray*}
~ p (l_{ijk} = {\rm dk}|R_l) = 1 - p (l_{ijk} = {\rm yes}|R_l) - p(l_{ijk} = {\rm no}|R_l)
\end{eqnarray*}

The remaining question is how to estimate $p(y_h|R_l)$, $h \in \{i, j, k\}$ for triplet
$(\mathbf{x}_i, \mathbf{x}_j, \mathbf{x}_k)$. Distance metric learning algorithms learn a distance
metric that is expected to reflect the relationship among the data points in the triplets. While the learned metric
does not directly provide us with a definition of the underlying classes, we could estimate the classes using the
learned distances via a variety of methods. One such method is clustering. The principal motivation behind using
clustering in this case is that the instances placed in the same cluster are very likely to come from the same
class since the distance metric is learned based on relative comparisons generated from the class labels. We summarize
this procedure as follows:
\begin{itemize}
\label{item:ranForest}
	\item Learn a distance metric for $\mathcal{D}$ with labeled triplets $R_l$.
	\item Use a clustering algorithm to partition the data $\mathcal{D}$ into $C$ clusters based on the learned metric.
	\item View the cluster assignments as a surrogate for the true class labels, and build a predictive model to estimate the class probability of each instance.
\end{itemize}

We remark that different methods can be used for this general procedure. In this paper, we use an existing metric learning algorithm by Schultz and Joachims~\cite{schultz2003learning} for the first step. For the clustering step, we choose to employ the $k$-means algorithm due to its simplicity. Finally, for the last step, we build a random forest of 50 decision trees to predict the cluster labels given the instances and estimate $p(y_h|R_l)$ for instance ${\mb x}_h$ using all decision trees in the forest that are not directly trained on ${\mb x}_h$.

\subsection{Complexity Analysis}

In this section we analyze the run-time of our proposed algorithm. Lines 1 and 2 of our algorithm performs the initialization and random sampling of the pool, which takes $O(n)$ time if $|R_p|$ is set to $pn$ where $p$ is a constant (in our experiment, we set $p$ as $100$). In line 3, we perform metric learning, apply $k$-means with the learned distance metric and build the random forest (RF) classifier to estimate the probabilities. The complexity of metric learning varies depending on which distance metric learning method is applied. In our experiments we use the method in \cite{schultz2003learning}, which casts metric learning as a quadratic programming problem. This can be solved in polynomial time on the number of dimensions and linearly on $n$. Running the $k$-means method takes $O(n)$ time assuming fixed number of iterations (we ignore other constant factors introduced by $k$ and the feature dimension). Building the RF takes $O(N_T n \log n)$, where $N_T$ is the number of decision trees in RF and $n$ is the number of instances \cite{breiman2001random}. Lines 5 to 10 evaluate each unlabeled triplet in $R_p$ and select the best one to query it, which takes $O(n)$ time. Putting this together, the running time of our algorithm for selecting a single query is dominated by $O(N_T n\log n)$, plus the running time of the metric learning method chosen. In our experiments, we use warm start in the metric learning process, which employs the learned metric from the previous iteration to initialize the metric learning in the current iteration. This significantly reduces the observed run-time for metric learning and makes it possible to select a query in real time. 

\begin{figure*}[ht!]
\begin{center}
\begin{tabular}{cc}
Breast & Parkinson\\
\includegraphics[height=.20\textheight, width = .40\textwidth]{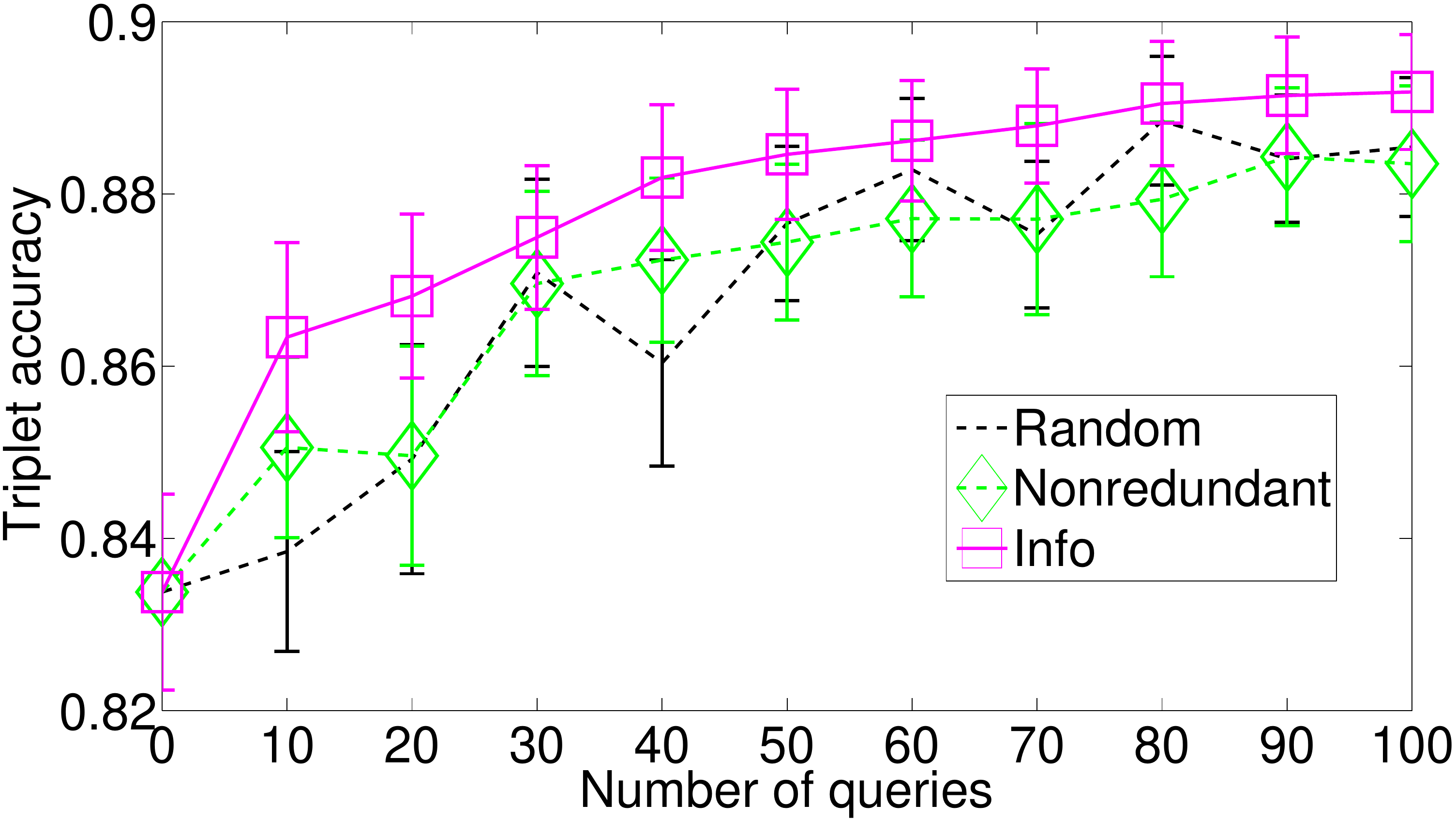}  &
\includegraphics[height=.20\textheight, width = .40\textwidth]{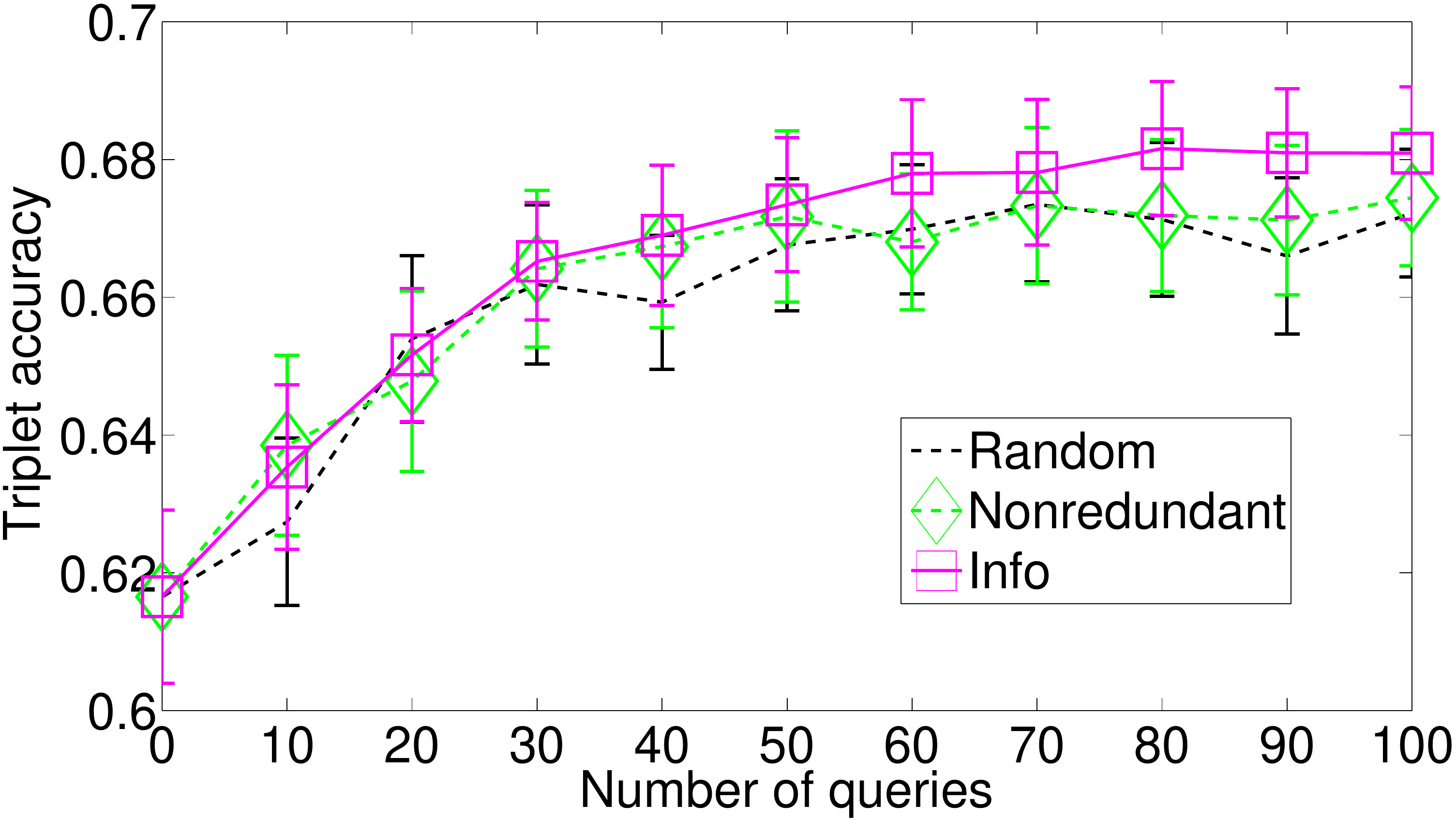} \\\\
Segment & Soybean \\
\includegraphics[height=.20\textheight, width = .40\textwidth]{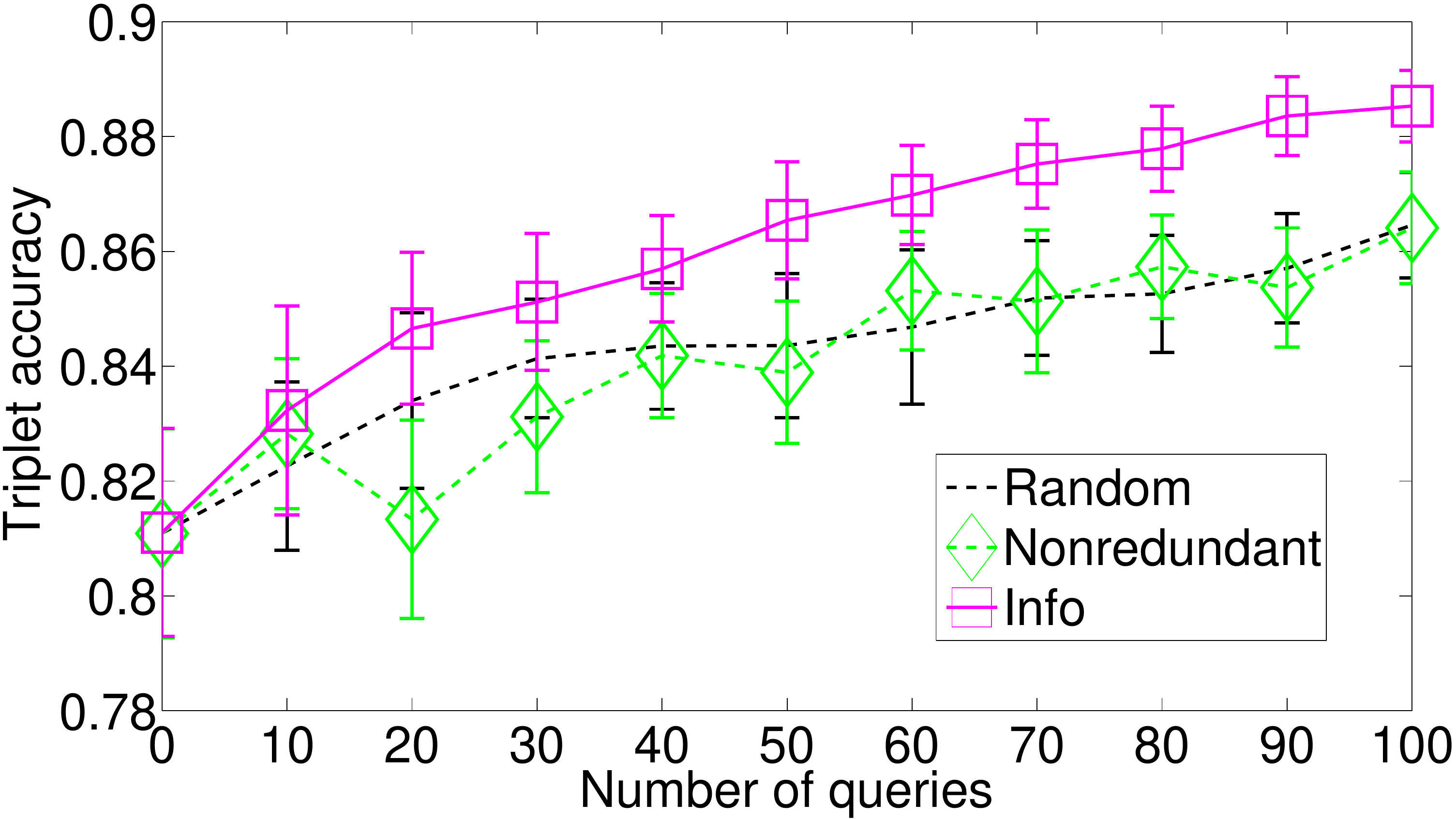} &
\includegraphics[height=.20\textheight, width = .40\textwidth]{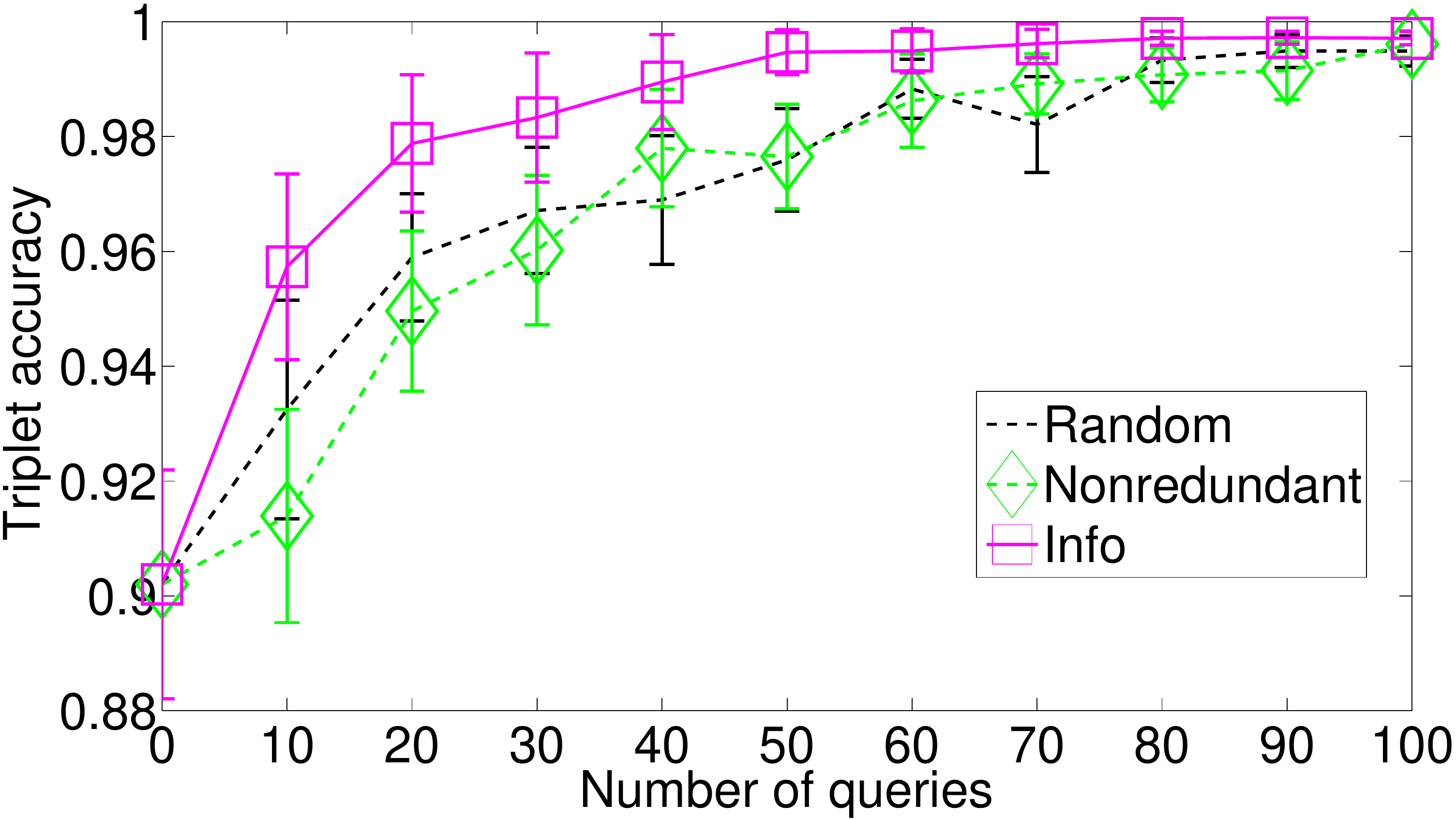}  \\\\
Wave & Wine \\
\includegraphics[height=.20\textheight, width = .40\textwidth]{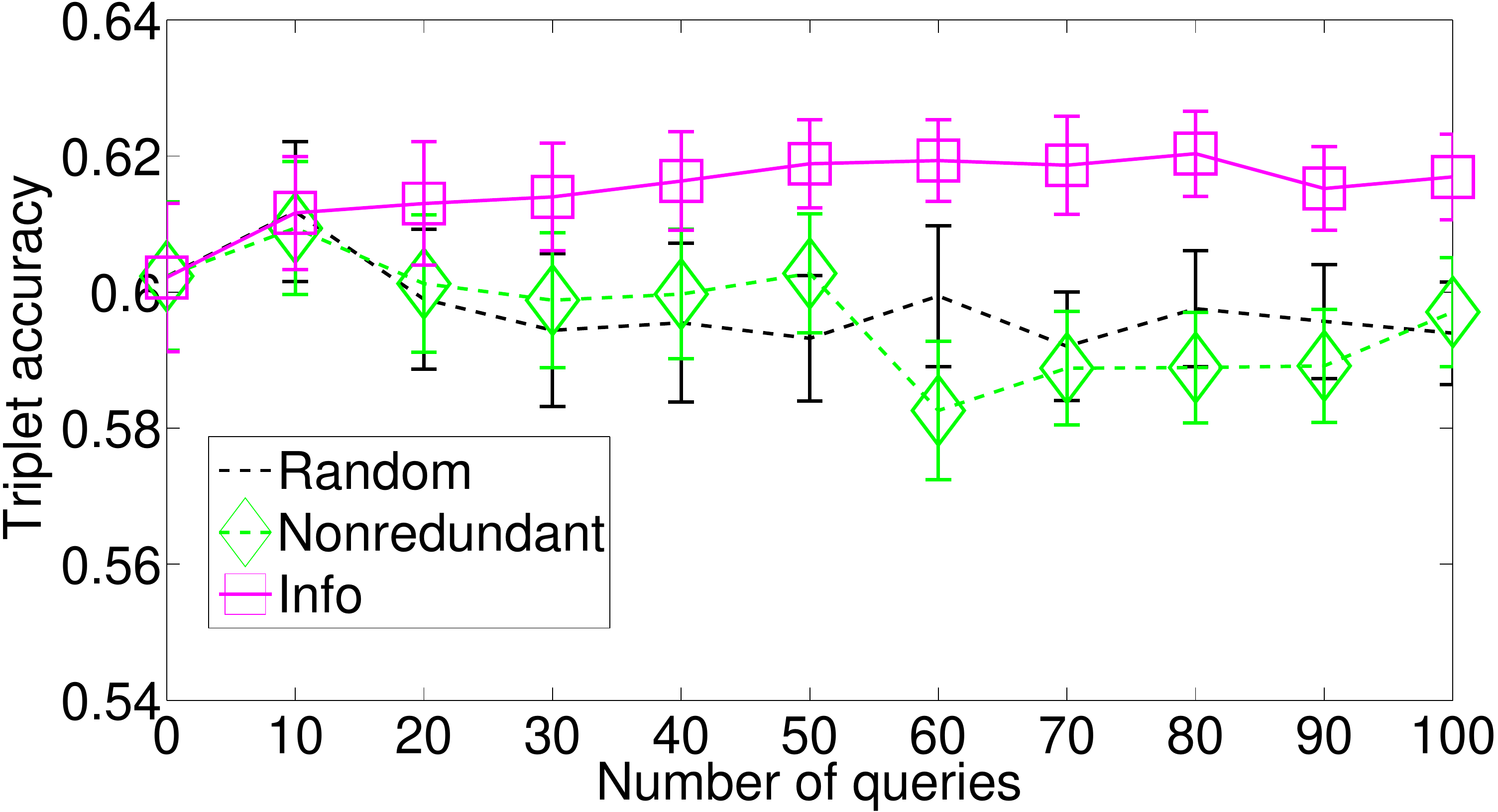} &
\includegraphics[height=.20\textheight, width = .40\textwidth]{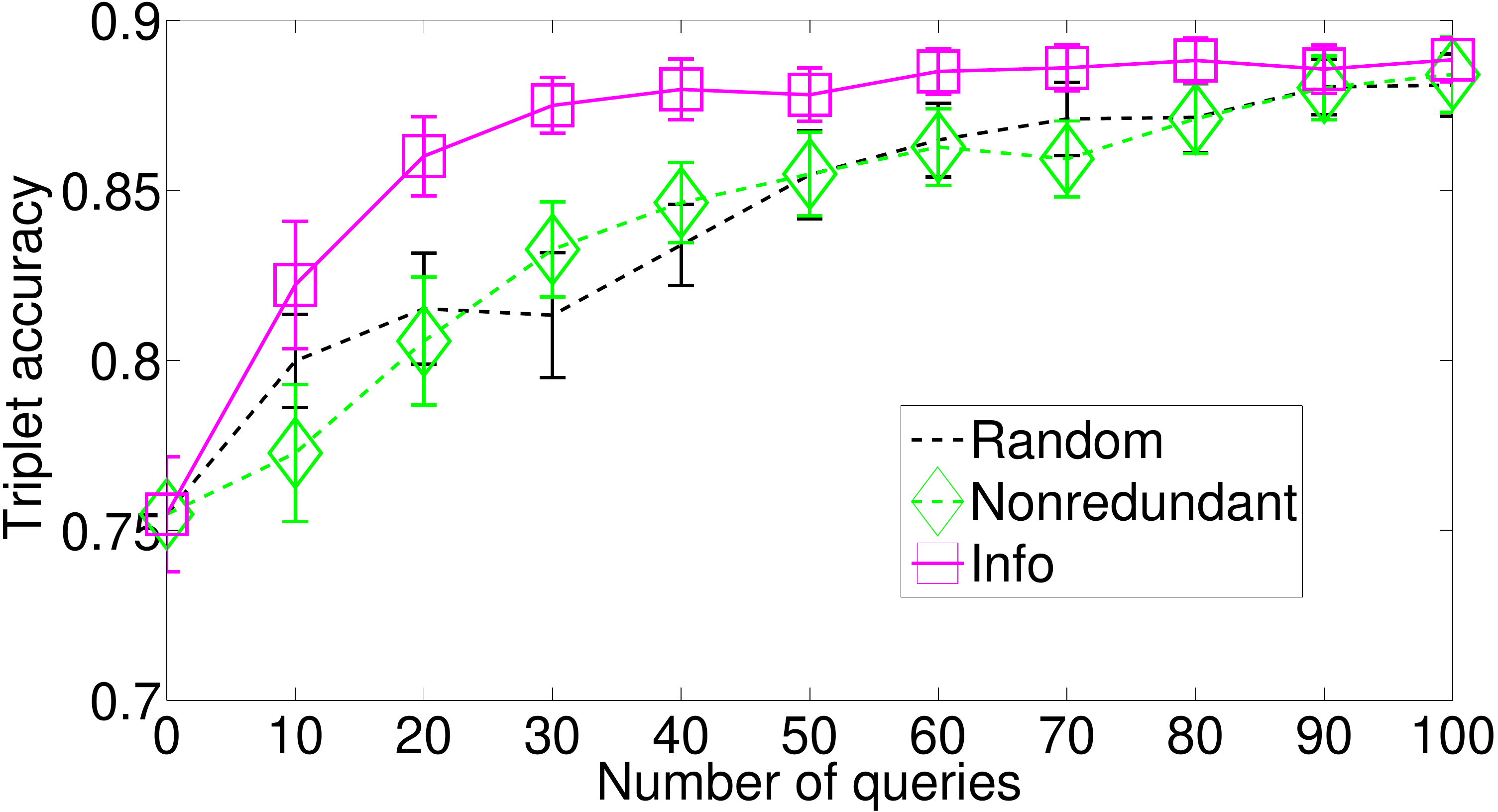}\\\\
Yeast  & HJA Birdsong\\
\includegraphics[height=.20\textheight, width = .40\textwidth]{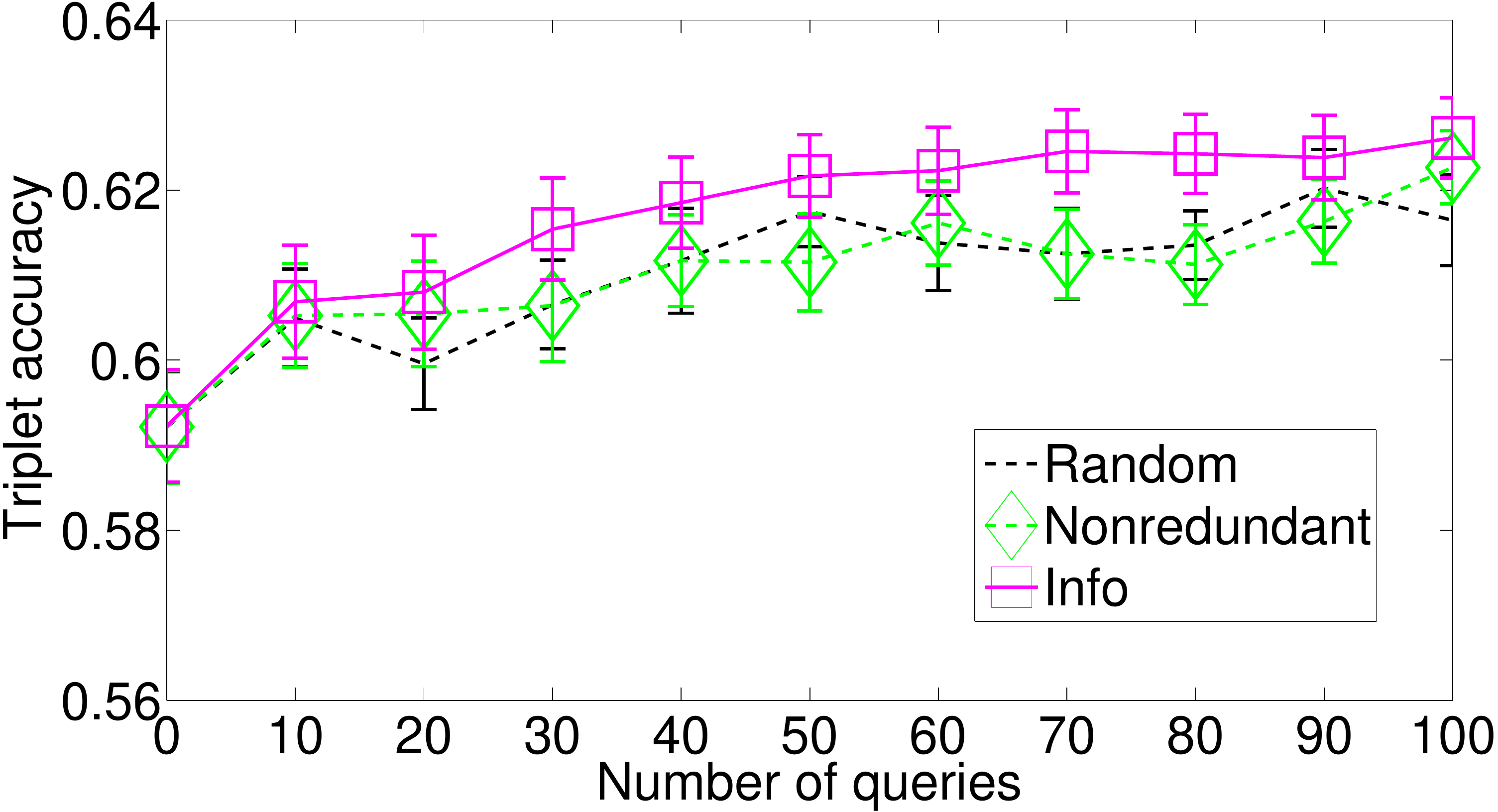}  &
\includegraphics[height=.20\textheight, width = .40\textwidth]{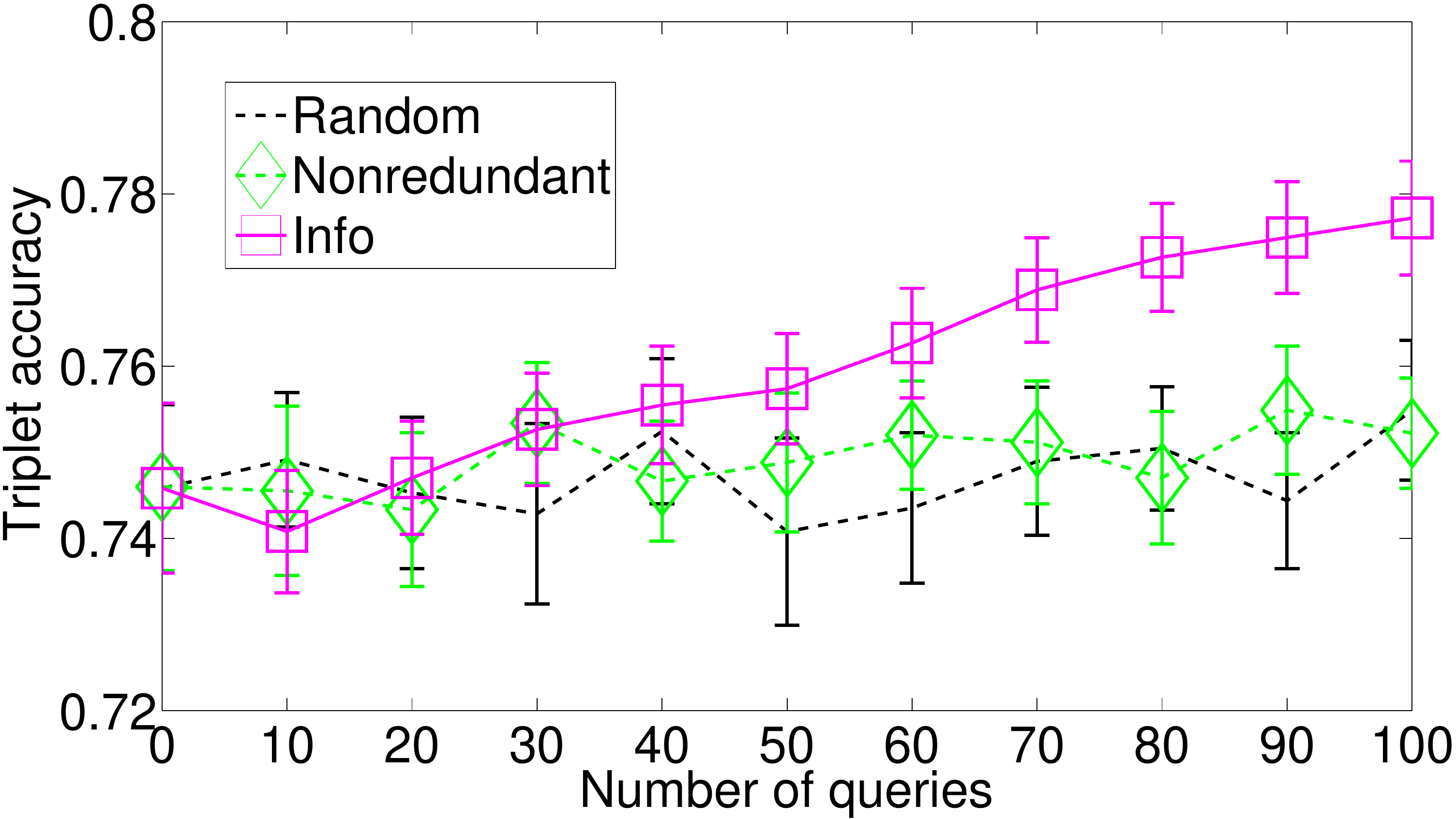}
\end{tabular}
\end{center}
\caption{Triplet accuracy for different methods as a function of the number of queries (error bars are shown as mean and $95\%$ confidence interval).}
\label{fig:results}
\end{figure*}

\section{Experiments}
\label{sec:expe}
In this section we experimentally evaluate the proposed method, which we will refer to as \emph{Info}.
Below we first describe the experimental setup for our evaluation and then present and discuss the experimental results.

\subsection{Datasets}
We evaluate our proposed method on seven benchmark datasets from the UCI Machine Learning repository~\cite{Frank+Asuncion:2010} and one additional larger datatset. The benchmark datasets include Breast Tissue (referred to as {\it Breast}), Parksinons~\cite{little2007exploiting}, Statlog image segmentation (referred to as {\it Segment} ), Soybean small, Waveform Database Generator (V.2) (referred to as {\it Wave}), Wine and Yeast. We also use an additional real-world dataset called HJA Birdsong. This dataset contains audio segments extracted from the spectrums of a collection of 10-second bird song recordings and uses the bird species as class labels\cite{briggs2012kdd}.

\begin{table}[ht!]
\caption{Summary of Datasets Information }
\label{tab:datasets}
\begin{center}
\begin{tabular}{c|c|c|c}
\hline
Name & \# Features & \# Instances  & \# Classes \\
\hline
Breast & 9 & 106 & 4 \\
Parkinson & 22 & 195 & 2 \\
Segment & 19 & 2310 & 7 \\
Soybean & 35 & 47 & 4 \\
Wave & 40 & 5000 & 2 \\
Wine & 13 & 178 & 3 \\
Yeast & 8 & 1484 & 9 \\
HJA Birdsong & 38 & 4998 & 13 \\
\hline
\end{tabular}
\end{center}
\end{table}

\begin{table*}[ht!]
\caption{Comparison on 1NN classification accuracy. The best/better method based on paired \emph{t}-tests at $p=0.05$ are highlighted in boldface.}
\label{tab:1NN}
\begin{center}
\begin{tabular}{cccccccccccc}
\hline
\multirow{2}{*}{Dataset} & \multirow{2}{*}{Algorithm} & \multicolumn{7}{c}{Number of queries} & \multirow{2}{*}{Prop. y/n answers}\\
\cline{3-9} & & 0 & 10 & 20 & 40 & 60 & 80 & 100 & \\
\hline
 \multirow{3}{*}{Breast}&Random&$\multirow{3}{*}{0.815}$&$0.816$&$0.832$&$0.849$&$\mathbf{0.857}$&$0.852$&$0.856$&$0.396$\\
&Nonredundant&&$0.828$&$0.830$&$0.846$&$0.845$&$0.851$&$\mathbf{0.863}$&$0.391$\\
&Info&&$\mathbf{0.839}$&$\mathbf{0.840}$&$\mathbf{0.858}$&$\mathbf{0.856}$&$\mathbf{0.861}$&$\mathbf{0.861}$&$\mathbf{0.558}$\\
\hline
 \multirow{3}{*}{Parkinson}&Random&$\multirow{3}{*}{0.817}$&${0.852}$&$\mathbf{0.863}$&$\mathbf{0.869}$&${0.867}$&${0.864}$&${0.865}$&${0.379}$\\
&Nonredundant&&${0.848}$&${0.858}$&${0.863}$&${0.869}$&${0.869}$&$\mathbf{0.870}$&${0.384}$\\
&Info&&${0.851}$&$\mathbf{0.866}$&$\mathbf{0.870}$&${0.869}$&${0.868}$&$\mathbf{0.872}$&$\mathbf{0.457}$\\
\hline
 \multirow{3}{*}{Segment}&Random&$\multirow{3}{*}{0.914}$&${0.923}$&${0.919}$&${0.920}$&$0.925$&$0.927$&$0.936$&$0.263$\\
&Nonredundant&&$\mathbf{0.932}$&${0.919}$&${0.919}$&$0.929$&$0.929$&$0.932$&$0.240$\\
&Info&&${0.924}$&${0.923}$&$\mathbf{0.933}$&$\mathbf{0.942}$&$\mathbf{0.947}$&$\mathbf{0.955}$&$\mathbf{0.394}$\\
\hline
 \multirow{3}{*}{Soybean}&Random&$\multirow{3}{*}{0.962}$&$0.981$&$0.987$&$0.989$&$0.990$&$0.991$&$0.991$&$0.338$\\
&Nonredundant&&$0.963$&$\mathbf{0.990}$&$\mathbf{0.991}$&$\mathbf{0.993}$&$\mathbf{0.992}$&$\mathbf{0.993}$&$0.334$\\
&Info&&$\mathbf{0.989}$&$\mathbf{0.995}$&$\mathbf{0.996}$&$\mathbf{0.997}$&$\mathbf{0.998}$&$\mathbf{0.999}$&$\mathbf{0.950}$\\
\hline
 \multirow{3}{*}{Wave}&Random&$\multirow{3}{*}{0.748}$&$0.766$&$0.747$&$0.732$&$0.741$&$0.744$&$0.742$&${0.470}$\\
&Nonredundant&&$0.756$&$0.743$&$0.746$&$0.730$&$0.736$&$0.746$&$0.437$\\
&Info&&$\mathbf{0.776}$&$\mathbf{0.776}$&$\mathbf{0.773}$&$\mathbf{0.772}$&$\mathbf{0.774}$&$\mathbf{0.779}$&$\mathbf{0.509}$\\
\hline
 \multirow{3}{*}{Wine}&Random&$\multirow{3}{*}{0.768}$&$0.853$&$0.878$&$0.922$&$0.943$&$0.948$&$0.949$&$0.434$\\
&Nonredundant&&$0.826$&$0.875$&$0.934$&$0.946$&$0.950$&$\mathbf{0.956}$&$0.449$\\
&Info&&$\mathbf{0.903}$&$\mathbf{0.946}$&$\mathbf{0.951}$&$\mathbf{0.954}$&$\mathbf{0.958}$&$\mathbf{0.959}$&$\mathbf{0.925}$\\
\hline
 \multirow{3}{*}{Yeast}&Random&$\multirow{3}{*}{0.403}$&$0.433$&$0.427$&$0.446$&$0.455$&$0.458$&$0.459$&${0.321}$\\
&Nonredundant&&$\mathbf{0.443}$&$\mathbf{0.434}$&$\mathbf{0.452}$&$0.455$&$0.458$&$\mathbf{0.464}$&${0.320}$\\
&Info&&$\mathbf{0.441}$&$\mathbf{0.434}$&$\mathbf{0.457}$&$\mathbf{0.467}$&$\mathbf{0.469}$&$\mathbf{0.470}$&$\mathbf{0.384}$\\
\hline
 \multirow{3}{*}{HJA Birdsong}&Random&$\multirow{3}{*}{0.660}$&${0.672}$&$0.659$&$0.661$&$0.650$&$0.640$&$0.647$&$0.231$\\
&Nonredundant&&${0.671}$&$\mathbf{0.671}$&$0.657$&$0.650$&$0.651$&$0.642$&$0.220$\\
&Info&&${0.675}$&$\mathbf{0.676}$&$\mathbf{0.681}$&$\mathbf{0.688}$&$\mathbf{0.694}$&$\mathbf{0.697}$&$\mathbf{0.407}$\\
\hline
\end{tabular}
\end{center}
\end{table*}

\subsection{Baselines and Experimental Setup}
\begin{table*}[ht!]
\caption{Win/tie/loss counts of Info versus baselines with varied numbers of queries based on 1NN classification accuracy.}
\label{tab:wtl}
\begin{center}
\begin{tabular}{cccccccc}
\hline
\multirow{2}{*}{Algorithms} & \multicolumn{6}{c}{Number of queries} & \multirow{2}{*}{In All}\\
\cline{2-7} & 10 & 20 & 40 & 60 & 80 & 100\\
\hline
Random & 5/3/0 & 6/2/0 & 7/1/0 & 6/2/0 & 7/1/0 & 8/0/0 & 39/9/0 \\
Nonredundant & 4/3/1 & 3/5/0 & 6/2/0 & 6/2/0 & 6/2/0 & 3/5/0 & 28/19/1\\
\hline
In All & 9/6/1 & 9/7/0 & 13/3/0 & 12/4/0 & 13/3/0 & 11/5/0 & 67/28/1 \\
\hline
\end{tabular}
\end{center}
\end{table*}

To the best of our knowledge, there is no existing active learning algorithm for relative comparisons.
To investigate the effectiveness of our method, we compare it with several randomized baseline
policies\footnote{We do not compare against active learning of pairwise constraints because the
differences in query forms and the requirement of different metric learning algorithms prevents us
from drawing direct comparisons.}:
\begin{itemize}
 \item {\it Random}: in each iteration, the learner randomly selects an unlabeled triplet to query.
 \item {\it Nonredundant}: in each iteration, the learner selects the unlabeled triplet that has
 the least instance overlap with previously selected triplets. If multiple choices exist, randomly choose one.
\end{itemize}

We use the distance metric learning algorithm introduced by \cite{schultz2003learning}. This algorithm formulates a constrained optimization problem where the constraints are defined by the triplets and aims at learning a distance metric that remains as close to an unweighted Euclidean metric as possible. We have also considered an alternative metric learning method introduced by \cite{rosales2006learning} obtaining consistent results.


Given a triplet query $(\mathbf{x}_i, \mathbf{x}_j, \mathbf{x}_k)$, the oracle returns an answer based on the unknown class labels $y_i, y_j$, and $y_k$. Since the datasets above provide class labels (but no triplet relationships), for these experiments we let the oracle return query answer yes if $y_i=y_j\neq y_k$, and no if $y_i=y_k\neq y_j$. In all other cases, the oracle returns dk. Both our method and the baseline policies may select triplets whose answer is dk. Such triplets cannot be utilized by the metric learning algorithm, but are counted as used queries since we are evaluating the active learning method (not the metric learning method as this is kept fixed across all the experiments).

In all the experiments, we randomly split the dataset $\mathcal{D}$
into two folds, one fold for training and the other for testing. We
initialize the active learning process with two randomly chosen yes/no
triplets (and the metric learning procedure with the identity matrix),
and then iteratively select one query at a step, up to a total of
$100$ queries. All query selection methods are initialized with the
same initial triplets. This process is repeated for $50$ independent
random runs and the reported results are averaged across the $50$
runs.

\begin{table*}[ht!]\small
\caption{Comparison on 1NN classification accuracy with varying number of {yes/no} triplets. The best/better performance based on paired \emph{t}-tests at $p=0.05$ are highlighted in boldface.}
\label{tab:1NN-2}
\begin{center}
\begin{tabular}{c||c|cccc||c|ccccc}
\hline
 \multirow{2}{*}{Algorithm} & \multirow{2}{*}{Dataset}  & \multicolumn{4}{c||}{Number of {yes/no} triplets}
 & \multirow{2}{*}{Dataset}  & \multicolumn{4}{c}{Number of {yes/no} triplets}\\
\cline{3-6} \cline{8-11}&  & 5 & 10 & 15 & 20 & & 5 & 10 & 15 & 20 \\
\hline
Random        &\multirow{3}{*}{Breast}    &$0.821$&$0.829$&$0.827$&$0.831$&\multirow{3}{*}{Wave} &$0.763$&$0.749$&$0.757$&$0.745$\\
Nonredundant  &                           &$0.826$&$0.834$&$0.840$&$0.841$&                      &$0.765$&$0.746$&$0.746$&$0.735$\\
Info &&$\mathbf{0.844}$&$\mathbf{0.849}$&$\mathbf{0.852}$&$\mathbf{0.858}$&                      &$0.773$&$\mathbf{0.777}$&$\mathbf{0.779}$&$\mathbf{0.775}$\\
\hline
Random      & \multirow{3}{*}{Parkinson} &$0.850$&$0.870$&$0.869$&$0.868$&\multirow{3}{*}{Wine}&$0.868$&$0.894$&$0.923$&$0.940$\\
Nonredundant&                             &$0.859$&$0.869$&$0.868$&$0.866$&                     &$0.857$&$0.882$&$0.924$&$0.932$\\
Info        &                             &$0.851$&$0.871$&$0.865$&$0.873$&                     &$\mathbf{0.895}$&$\mathbf{0.930}$&$\mathbf{0.939}$&$0.943$\\
\hline
Random      & \multirow{3}{*}{Segment}      &$0.903$&$0.912$&$0.914$&$0.921$& \multirow{3}{*}{Yeast} &$0.409$&$0.415$&$0.419$&$0.423$\\
Nonredundant&                               &$0.909$&$0.913$&$0.905$&$0.923$&                        &$0.423$&$0.428$&$0.434$&$0.440$\\
Info  & &$\mathbf{0.921}$&$\mathbf{0.928}$&$\mathbf{0.934}$&$\mathbf{0.940}$&                        &$\mathbf{0.437}$&$\mathbf{0.448}$&$\mathbf{0.456}$&$\mathbf{0.460}$\\
\hline
Random      &\multirow{3}{*}{Soybean}       &$0.978$&$0.990$&$0.993$&$0.994$&\multirow{3}{*}{HJA Birdsong} &$0.672$&$0.669$&$0.668$&$0.662$\\
Nonredundant&                               &$0.979$&$0.988$&$0.996$&$0.995$&                              &$0.666$&$0.650$&$0.648$&$0.659$\\
Info        &                      &$\mathbf{0.988}$&$0.991$&$0.993$&$0.994$&                              &$0.671$&$\mathbf{0.680}$&$\mathbf{0.684}$&$\mathbf{0.686}$\\
\hline
\end{tabular}
\end{center}
\end{table*}

\begin{figure*}[t!]
\begin{center}
\begin{tabular}{ccc}
Breast & Soybean & Wine \\
\includegraphics[height=.20\textheight, width = .32\textwidth]{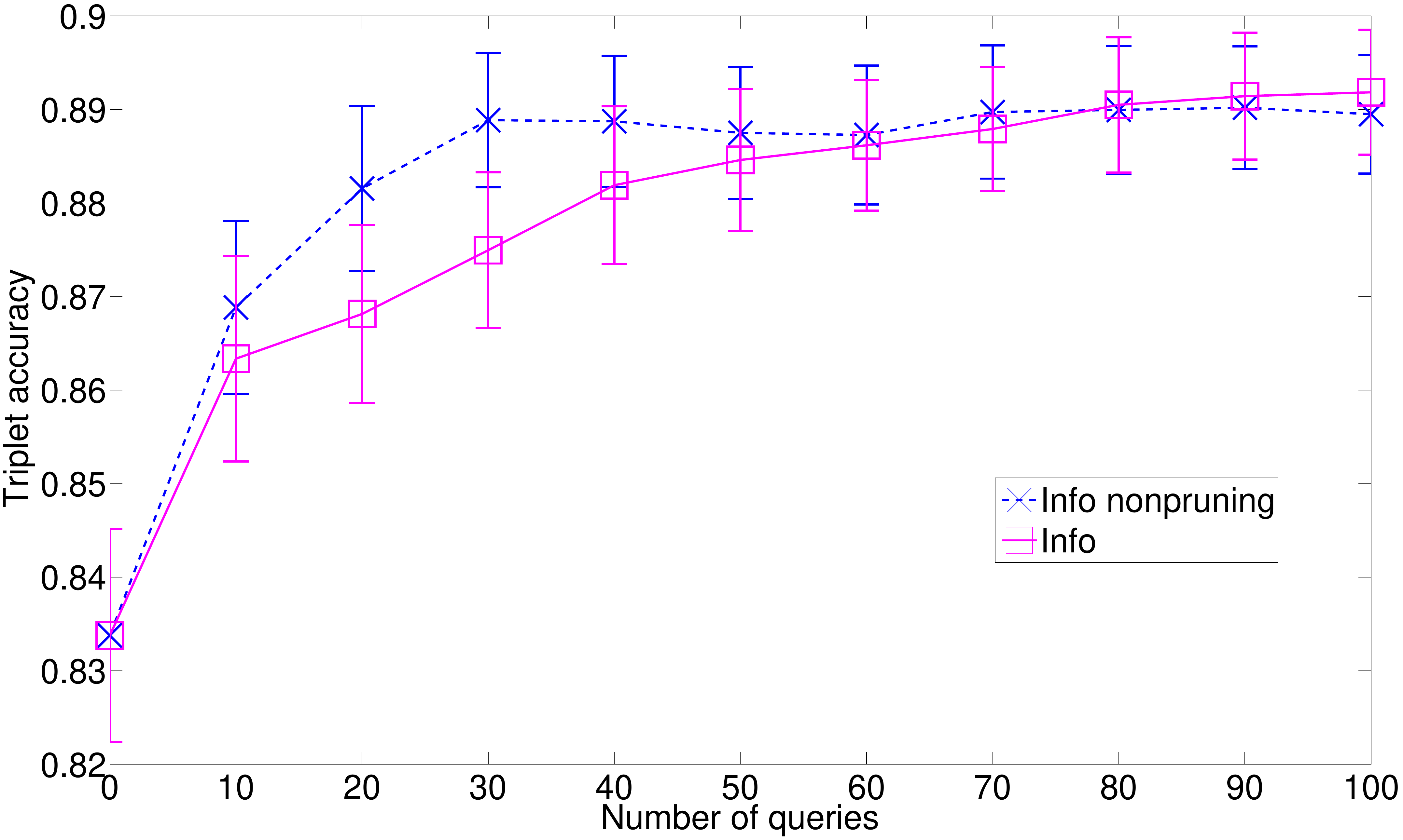}  &
\includegraphics[height=.20\textheight, width = .32\textwidth]{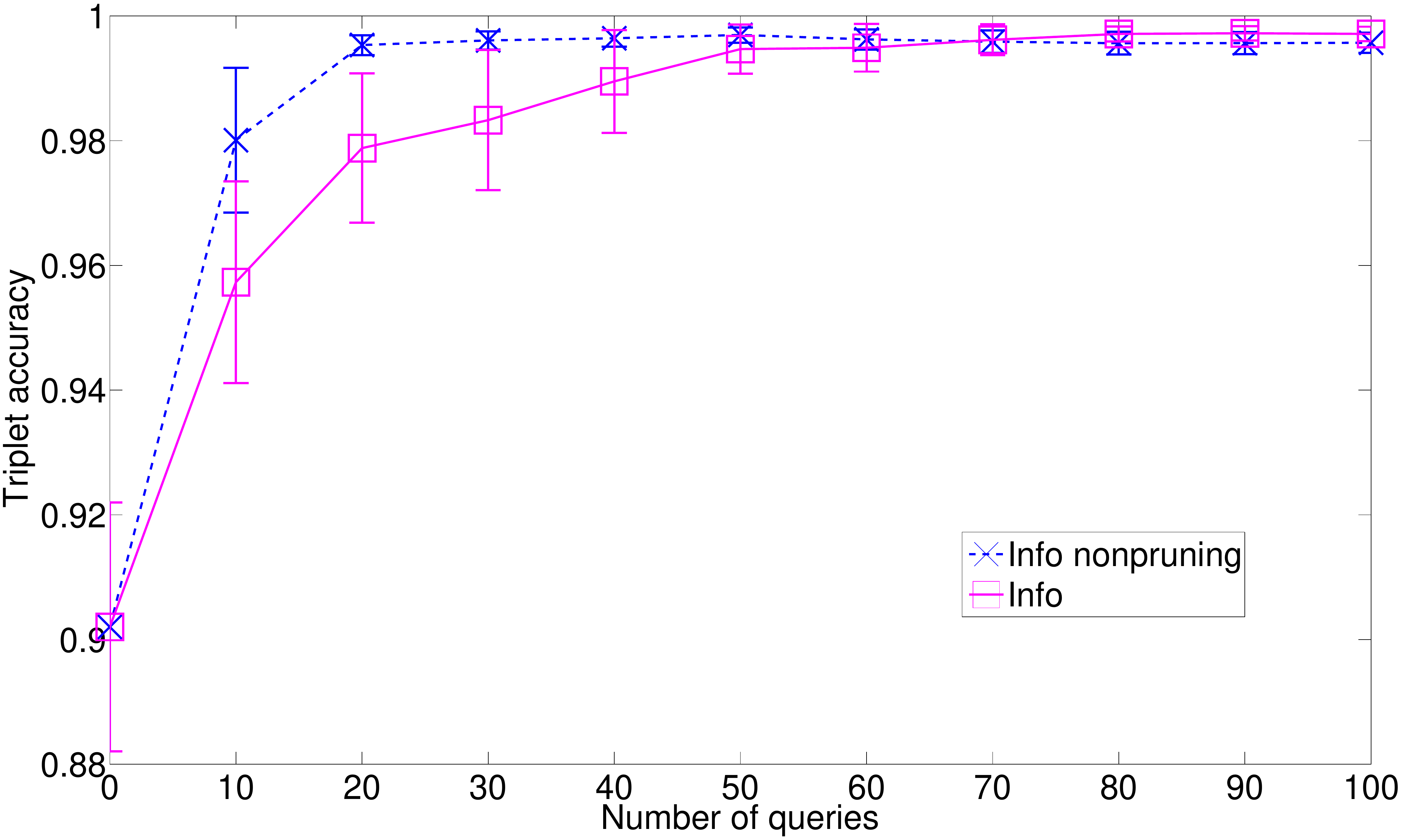} &
\includegraphics[height=.20\textheight, width = .32\textwidth]{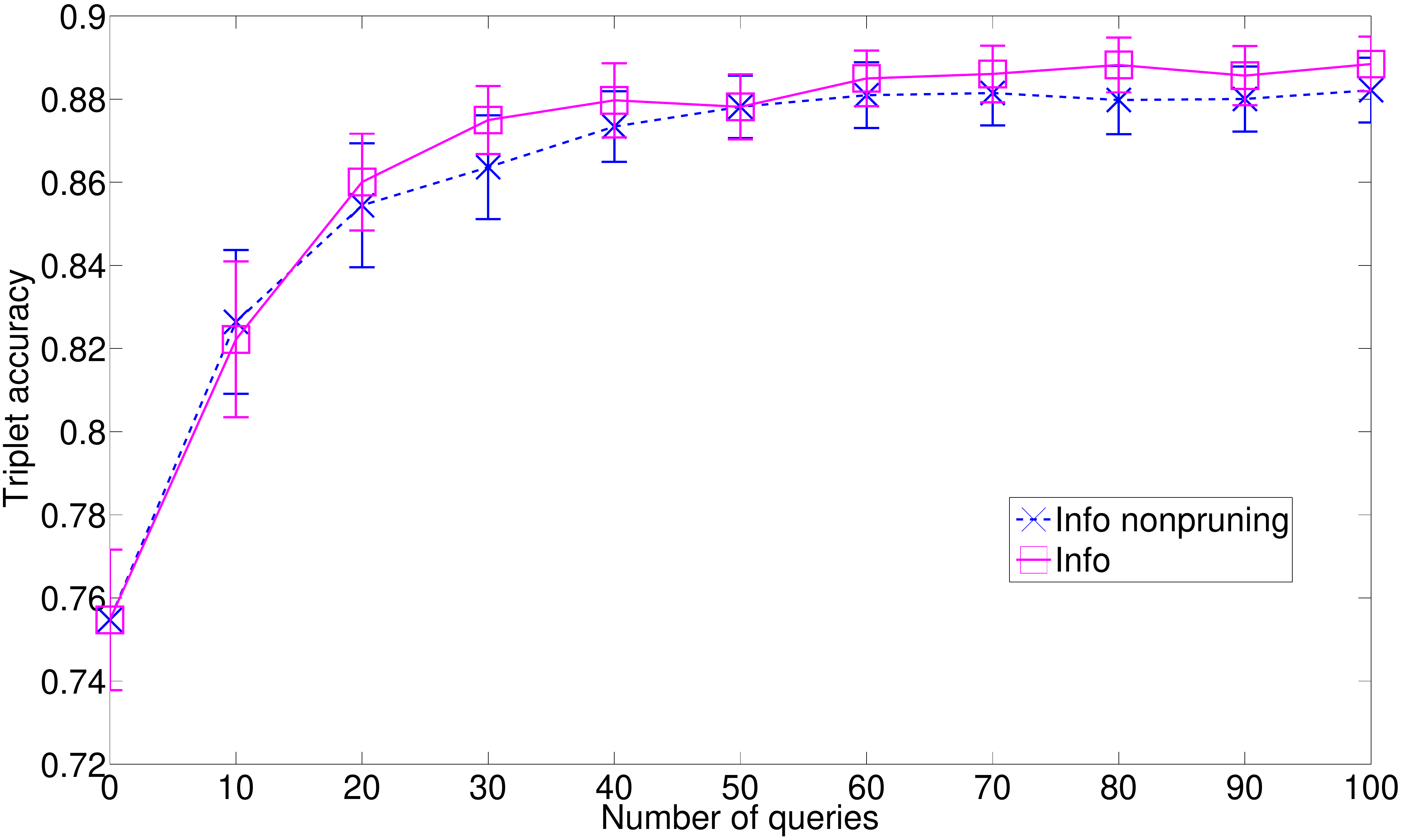} \\\\
\end{tabular}
\end{center}
\vspace{-0.5cm}
\caption{Triplet accuracy for the exact Info method and the approximation based on the sampling procedure (Sec.~\ref{sec:prune}) as a function of the number of queries (error bars are shown as mean and $95\%$ confidence interval).}
\label{fig:prune}
\end{figure*}

\subsection{Evaluation Criteria}
We evaluate the learned distance metric using two performance measures. The first is triplet accuracy,
which evaluates how accurately the learned distance metric can predict the answer to an unlabeled {yes/no}
triplet from the test data. This is a common criterion used in previous studies on metric learning with relative
comparisons~\cite{rosales2006learning,schultz2003learning}. To create the triplet testing set, we generate all
triplets with {yes/no} labels from the test data for small datasets including Breast, Parkinson, Soybean, Wine.
For larger datasets, we randomly sample $200$K {yes/no} triplets from the test set to estimate the triplet accuracy.

For the second measure, we record the classification accuracy of a $1$-nearest neighbor (1NN) classifier \cite{1NN} on the test data based on the learned distance metric. The purpose of this measurement is to examine how much the learned distance metric can improve the classification accuracy.

\subsection{Results}


Fig.~\ref{fig:results} plots the triplet accuracy obtained by our
method (denoted by {\it Info}) and the baseline methods as a function
of the number of queries ranging from $0$ to $100$. The results are
averaged over 50 random runs.  Table \ref{tab:1NN} shows the 1NN
classification accuracy, with 10, 20, 40, 60, 80, and 100 queries,
respectively.  We also list the 1NN accuracy without any queries as
the initial performance. For each case, the best result(s) is/are
highlighted in boldface based on paired \emph{t}-tests at
$p=0.05$. The win/tie/loss result (from Table \ref{tab:1NN}) for Info
against each method separately is summarized in Table \ref{tab:wtl}.

First, we observe that Random and Nonredundant have a decent performance in most datasets and the two baselines
do not differ significantly. 
These results are consistent with what has been reported in previous metric-learning studies where the learned metric tends to improve as we include more and more randomly selected yes/no triplets. For the two large datasets, Wave and HJA Birdsong, Random and Nonredundant do not perform well. A possible explanation for this is the large number of data points, which may require more triplets to learn a good metric.

We can also see that Info consistently outperforms Random and Nonredundant. In particular, we see that for some datasets (e.g., Segment, Wave, Yeast and HJA Birdsong), Info was able to achieve better triplet accuracy than the two random baselines. For some other datasets (e.g.,  Soybean and Wine), Info achieved the same level of performance but with significantly fewer queries than the two baseline methods. Also for these datasets, the variance of Info is generally smaller (as indicated by its smaller confidence intervals). For the Breast and Parkinson datasets, the performance of Info does not significantly outperform the two baseline methods, but generally tends to produce smoother curves where the baseline methods appear inconsistent as we increase the number of queries. Overall the results suggest that the triplets selected by Info are more informative than those selected randomly.

Similar behavior can also be observed from Table \ref{tab:1NN}  where the distance metric is evaluated using 1NN classification accuracy. In particular, we observe that Info often improves the performance over Random, and only suffers one loss to Nonredundant when the query number is small in the Segment dataset.

\subsection{Further Investigation}
Below we examine various important factors that contribute to the success of Info. Recall that all methods could produce queries of triplets labeled as dk (don't know), yet the distance metric learning algorithm can only learn from yes/no triplets. Info explicitly considers this effect by setting $H(y_{ijk}|l_{ijk}={dk}, R_l) = H(y_{ijk}|R_l)$.
Thus we expect Info to avoid queries that are likely to return a dk answer, which could be an important factor contributing to its
winning over the baselines. We test this hypothesis by recording the percentage of selected triplets with yes/no answers in $100$
queries averaged by $50$ runs, which are reported in the last column of Table \ref{tab:1NN}. From the results we observe that
the percentages vary significantly from one dataset to another. But Info typically achieves a much higher percentage compared
to the two baselines, which confirms our hypothesis.

Besides the first factor that Info can select more triplets of {yes/no} answers,
we would like to ask if the {yes/no} triplets selected by Info are more informative than those selected by the random baselines.
To answer this question, we examine the 1NN classification accuracies achieved by each method as a function of the number of {yes/no} triplets.
This will allow us to examine the usefulness of the final constraints obtained by each method.  We vary the number of yes/no triplets from 5 to
20 with an increment of 5 and then record the results of 50 independent random runs. The averaged 1NN classification accuracy results are reported in Table \ref{tab:1NN-2}. We can observe that Info still leads to better performance than the baselines,  suggesting that the triplets selected by Info are indeed more informative toward learning a good metric.


Finally, we want to examine the impact of the sampling approach described in Sec.~\ref{sec:prune} on our algorithm. To this end, we run our proposed algorithm without random sampling on three small datasets Breast, Soybean and Wine. Fig.~\ref{fig:prune} presents the performance of our exact algorithm (Info Exact) and with sampling (Info). It can be observed that sampling has some minor impact on the Breast and Soybean datasets, and no impact on the Wine dataset. This suggests that the probabilistic guarantees of the random sampling method are indeed sufficient in practice to avoid significant performance loss compared to the exact active learning version.


To summarize, the empirical results demonstrate that Info can consistently outperform the random baselines. In particular, we demonstrate that it has two advantages. First, it allows us to effectively avoid querying {dk} triplets, resulting in more yes/no triplets to be used by the metric learning algorithm.  Second, empirical results suggest that the {yes/no} triplets selected by our method are generally more informative than those selected by the random baselines. Finally, the results suggest that while sampling approximation may negatively impact the performance in some cases, the impact is very mild, both in theory and practice, and the resulting algorithm still consistently outperforms the random baselines.

\section{Conclusion}
\label{sec:concl}
This paper studied active distance metric learning from relative comparisons. In particular, it considered queries of the form: {\it is instance $i$ more similar to instance $j$ than to $k$?}. The approach utilized the existence of an underlying (but unknown) class structure. This class structure is implicitly employed by the user/oracle to answer such queries. In formulating active learning in this setting, we found of interest that some query answers cannot be utilized by existing metric learning algorithms: the answer {\it don't know} does not provide any useful constraint for these. The presented approach addressed this in a natural, sound manner. We proposed an information theoretic objective that explicitly measures how much information about the class labels can be obtained from the answer to a query. In addition, to reduce the complexity of selecting a query, we showed that a simple sampling scheme can provide excellent performance guarantees. Experimental results demonstrated that the triplets selected by the proposed method not only contributed to learning better distance metrics than those selected by the baselines, but helped improve the resulting classification accuracy.

\end{document}